\documentclass[9pt,conference]{IEEEtran}
\usepackage[T1]{fontenc}
\usepackage{lmodern}

\usepackage{cite}
\usepackage{amsmath,amssymb,amsfonts}
\usepackage{algorithmic}
\usepackage{graphicx}
\usepackage{textcomp}
\usepackage{xcolor}
\usepackage{booktabs}

\def\BibTeX{{\rm B\kern-.05em{\sc i\kern-.025em b}\kern-.08em
    T\kern-.1667em\lower.7ex\hbox{E}\kern-.125emX}}
\begin{document}

\title{PDB-Eval: An Evaluation of Large Multimodal Models for Description and Explanation of Personalized Driving Behavior
% \\
% {\footnotesize \textsuperscript{*}Note: Sub-titles are not captured for https://ieeexplore.ieee.org  and
}

% \author{\IEEEauthorblockN{1\textsuperscript{st} Junda Wu}
% % \IEEEauthorblockA{\textit{Computer Science and Engineering} \\
% \IEEEauthorblockA{\textit{Computer Science and Engineering} \\
% \textit{University of California San Diego}\\
% La Jolla, USA \\
% juw069@ucsd.edu}
% \and
% \IEEEauthorblockN{2\textsuperscript{nd} Jessica Echterhoff}
% \IEEEauthorblockA{\textit{Computer Science and Engineering} \\
% \textit{University of California San Diego}\\
% La Jolla, USA \\
% jechterh@ucsd.edu}
% \and
% % \IEEEauthorblockN{3\textsuperscript{rd} An Yan}
% % \IEEEauthorblockA{\textit{Computer Science and Engineering} \\
% % \textit{UCSD}\\
% % La Jolla, USA \\
% % ayan@eng.ucsd.edu}
% % \and
% \IEEEauthorblockN{3\textsuperscript{rd} Kyungtae Han}
% \IEEEauthorblockA{\textit{InfoTech Labs} \\
% \textit{Toyota Motor North America}\\
% Mountain View, USA \\
% kt.han@toyota.com}
% \and
% \IEEEauthorblockN{4\textsuperscript{th} Amr Abdelraouf}
% \IEEEauthorblockA{\textit{InfoTech Labs} \\
% \textit{Toyota Motor North America}\\
% Mountain View, USA \\
% amr.abdelraouf@toyota.com}
% \and
% \IEEEauthorblockN{5\textsuperscript{th} Rohit Gupta}
% \IEEEauthorblockA{\textit{InfoTech Labs} \\
% \textit{Toyota Motor North America}\\
% Mountain View, USA \\
% rohit.gupta@toyota.com}
% \and
% \IEEEauthorblockN{6\textsuperscript{th} Julian McAuley}
% \IEEEauthorblockA{\textit{Computer Science and Engineering} \\
% \textit{University of California San Diego}\\
% La Jolla, USA \\
% jmcauley@eng.ucsd.edu}
% }

\author{
\begin{tabular}{ccc}
\begin{tabular}{@{}c@{}}
Junda Wu \\
\textit{Computer Science and Engineering} \\
\textit{University of California San Diego} \\
La Jolla, USA \\
juw069@ucsd.edu
\end{tabular}
&
\begin{tabular}{@{}c@{}}
Jessica Echterhoff \\
\textit{Computer Science and Engineering} \\
\textit{University of California San Diego} \\
La Jolla, USA \\
jechterh@ucsd.edu
\end{tabular}
&
\begin{tabular}{@{}c@{}}
Kyungtae Han \\
\textit{InfoTech Labs} \\
\textit{Toyota Motor North America} \\
Mountain View, USA \\
kt.han@toyota.com
\end{tabular}
\\
\\
\begin{tabular}{@{}c@{}}
Amr Abdelraouf \\
\textit{InfoTech Labs} \\
\textit{Toyota Motor North America} \\
Mountain View, USA \\
amr.abdelraouf@toyota.com
\end{tabular}
&
\begin{tabular}{@{}c@{}}
Rohit Gupta \\
\textit{InfoTech Labs} \\
\textit{Toyota Motor North America} \\
Mountain View, USA \\
rohit.gupta@toyota.com
\end{tabular}
&
\begin{tabular}{@{}c@{}}
Julian McAuley \\
\textit{Computer Science and Engineering} \\
\textit{University of California San Diego} \\
La Jolla, USA \\
jmcauley@eng.ucsd.edu
\end{tabular}
\end{tabular}
}

\maketitle
\begin{abstract}
Understanding a driver's behavior and intentions is important for potential risk assessment and early accident prevention.
Safety and driver assistance systems can be tailored to individual drivers' behavior, significantly enhancing their effectiveness.
However, existing datasets are limited in describing and explaining general vehicle movements based on external visual evidence.
This paper introduces a benchmark, \textbf{PDB-Eval}, for a detailed understanding of \textbf{P}ersonalized \textbf{D}river \textbf{B}ehavior, 
and aligning Large Multimodal Models (MLLMs) with driving comprehension and reasoning.
Our benchmark consists of two main components, PDB-X and PDB-QA. 
PDB-X can evaluate MLLMs' understanding of temporal driving scenes. 
Our dataset is designed to find valid visual evidence from the external view to explain the driver's behavior from the internal view.
To align MLLMs' reasoning abilities with driving tasks, we propose PDB-QA as a visual explanation question-answering task for MLLM instruction fine-tuning.
As a generic learning task for generative models like MLLMs, PDB-QA can bridge the domain gap without harming MLLMs' generalizability.
Our evaluation indicates that fine-tuning MLLMs on fine-grained descriptions and explanations can effectively bridge the gap between MLLMs and the driving domain, 
which improves zero-shot performance on question-answering tasks by up to 73.2\%.
We further evaluate the MLLMs fine-tuned on PDB-X in Brain4Cars' intention prediction and AIDE's recognition tasks.
We observe up to 12.5\% performance improvements on the turn intention prediction task in Brain4Cars, 
and consistent performance improvements up to 11.0\% on all tasks in AIDE.
% The evaluation results showcase the effectiveness and generalizability of MLLMs' acquired descriptive and explaining abilities in downstream driving tasks.

% \keywords{MLLMs \and Personalized Driving Behavior \and Question-answering}
\end{abstract}


\section{Introduction} \label{sec:intro}
Driver understanding is critical for predicting vehicle movement \cite{jain2016brain4cars} 
and assessing potential risks on the road.
Recent research has shown advances in recognition tasks of traffic accidents, 
uncertainty,
and vehicle motion prediction \cite{yang2023aide}.
In such tasks, textual explanations for vehicle movement \cite{kim2018textual}
and driver behavior \cite{yang2023aide} 
have become increasingly important for more explainable understanding.
Since large multimodal models (MLLMs) can generate descriptions and explanations over visual evidence \cite{huang2023vtimellm, wu2024personalized, wu2024visual},
MLLMs have been recognized as a multimodal reasoner for driving tasks \cite{wu2024visual,li2023mvbench}.
However, existing MLLMs are limited to fine-tuning and evaluation on general visual understanding and explanation tasks \cite{li2023mvbench, ko2023large, yan2024list}, 
where a domain gap exists while adapting to driving tasks.

\begin{figure}
    \centering
    \includegraphics[width=.90\linewidth]{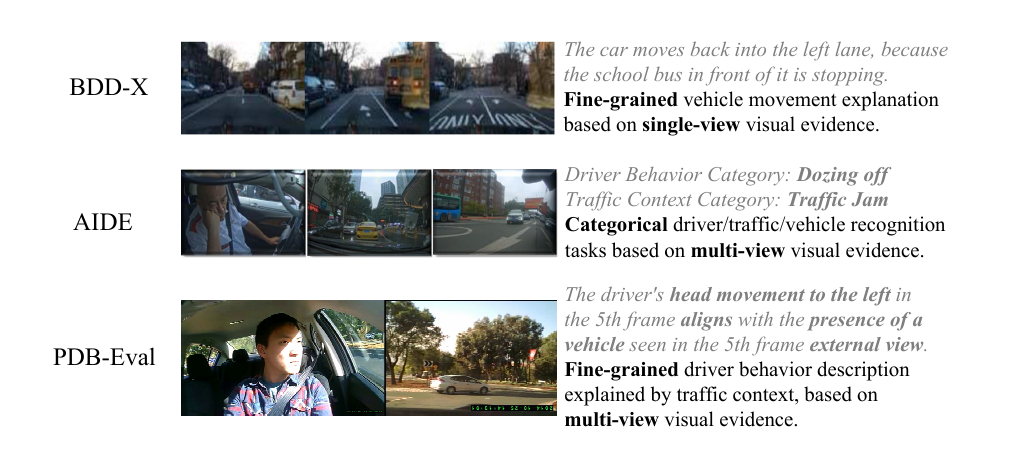}
    \caption{
        Existing vehicle movement understanding (\emph{e.g.}, BDD-X \cite{kim2018textual}) can be fine-grained but limited to single-view explanations.
        Multi-view and multi-task driving tasks (\emph{e.g.}, AIDE \cite{yang2023aide}) can provide multi-view visual evidence but lack description granularity.
        PDB-Eval can provide driver behavior descriptions explained by traffic context based on multi-view visual evidence.
        }
    \label{fig:demo}
    \vspace{-.5cm}
\end{figure}
To align MLLMs with driver understanding and reasoning tasks, we propose an evaluation dataset, \textbf{PDB-Eval}, for personalized driver behavior understanding.
While many existing works focus on analyzing driver behavior from the in-cabin view \cite{jain2016brain4cars,yang2023aide} or the external view of traffic scenes \cite{kim2018textual, bao2020uncertainty}, 
few have tried to understand these viewpoints as parallel time processes, 
where a video moment (\emph{i.e.}, an event happening between a time boundary) in one viewpoint can be explained based on the other viewpoint's visual evidence. 
For example, in Figure \ref{fig:demo}, existing fine-grained explanations of vehicle movements (\emph{e.g.}, BDD-X \cite{kim2018textual}) are limited to single-view understanding.
Multi-view driving understanding datasets (\emph{e.g.}, AIDE \cite{yang2023aide}) enable the understanding of multiple categories of driver behavior and vehicle movement, but they fall short in offering fine-grained descriptions.
To unify driving behavior description and explanation from both internal and external visual evidence, we introduce our first evaluation task \textbf{PDB-X}.
% This task focuses on explaining personalized driver behavior descriptions with corresponding visual explanations derived from external traffic scenes.
For example, in Figure \ref{fig:demo}, we show that PDB-X provides fine-grained driving behavior explanations based on multi-view visual evidence.
We further propose the visual explanation question-answering task \textbf{PDB-QA} to enhance the interpretative and reasoning capabilities of MLLMs in the context of driver behavior.
% PDB-QA is derived from PDB-X as a complex question-answering task that is compatible with general pipelines of MLLM instruction fine-tuning \cite{rajpurkar2016squad, lewis2019mlqa, brown2020language}.

Extracting personalized driver behavior descriptions can be challenging since human annotators must describe based on knowledge and observations of different types of drivers.
% In practice, we propose leveraging MLLMs as an intermediary step to generate preliminary versions of personalized driver behavior descriptions.
% Specifically, we introduce a comparative prompting method that prompts the MLLMs to generate only the driving behavior discrepancies between two drivers who have the same intention. 
% By focusing on discrepancies, the comparative prompting method ensures that the generated descriptions are specific and relevant to each driver's unique behavior.
We suggest using MLLMs to create personalized driver behavior descriptions by comparative prompting \cite{zhang2024cocot}, which highlights behavior discrepancies between drivers with the same intention, 
ensuring specificity and relevance in the descriptions.
Despite the potential of this approach, it is crucial to address the inherent challenge of hallucination in MLLM-generated content, 
where models may produce fabricated or irrelevant information \cite{ko2023large,alayrac2022flamingo}. 
This issue is particularly significant in tasks requiring detailed and accurate descriptions based on visual evidence \cite{ko2023large,alayrac2022flamingo}.
% To mitigate this risk, we first summarize the comparatively prompted answers into characteristic categories and the most prominent visual evidence corresponding to each type of driver.
% Using these categorized descriptions as guidelines, we can prompt the MLLMs again with more specific questions with fewer chances for the MLLMs to hallucinate irrelevant information.
To reduce this risk, we categorize prompted answers for each driver type, and then use these categories to guide more focused MLLM prompts, minimizing irrelevant information.

We evaluate two tasks in PDB-Eval by fine-tuning open-source MLLMs, BLIP-2 and VTimeLLM, which are specialized in image understanding and video understanding, respectively.
We also include GPT-4V as a strong zero-shot baseline to demonstrate the performance of the existing MLLMs.
In addition, we further evaluate the driver intention prediction tasks in the Brain4Cars dataset \cite{jain2016brain4cars} (in-domain) and driving recognition tasks in the AIDE dataset \cite{yang2023aide} (across-domain).
The evaluation results showcase the effectiveness and generalizability of MLLMs' acquired descriptive and explaining abilities via training on PDB-X.

% We summarize our contributions as follows:
% \begin{itemize}
%     \item We introduce the Personalized Driver Behavior Description and Explanation dataset (PDB-X), 
%     where the driver's behavior descriptions are explained based on visual evidence from the external view.
%     \item To extract contrasting behavior of different drivers, we introduce visual comparative prompting in MLLMs. 
%     \item We further introduce a visual explanation question-answering task (PDB-QA) that requires reasoning over internal and external visual evidence, 
%     challenging models to understand and explain driver behavior comprehensively.
%     \item We evaluate various MLLMs on PDB-X and PDB-QA based on their fine-tuned BLEU-4 performance.
%     We further evaluate the effectiveness of such acquired explanation ability in MLLMs by demonstrating improvements in prediction and recognition accuracy in various downstream tasks.
% \end{itemize}

\section{Related Work}

Recent driving understanding research spans tasks such as time-to-accident, intention, accident prediction, driving anticipation, distraction, and uncertainty estimation \cite{bao2020uncertainty, bonyani2023dipnet}. Despite progress, many systems lack fine-grained interpretability and reasoning, limiting trust in applications like self-driving cars. To address this, recent approaches incorporate textual explanations, question-answering based on dash-cam evidence \cite{kim2018textual, zhang2023study}, attention maps, and even concept bottleneck frameworks \cite{echterhoff2024driving} that verbalize drivers’ behavior. In contrast to static, categorical descriptions, our benchmark challenges models to understand personalized driver behavior and correlate it with dynamic vehicle and traffic conditions.

Large multimodal models (MLLMs) have been increasingly applied in driving-related tasks such as planning, navigation, simulation, and command understanding \cite{fu2024drive, xu2023drivegpt4}. However, while existing video-based MLLM evaluations primarily focus on single-view analysis, a key challenge remains: reasoning over two causally linked temporal processes—the internal dynamics of driver behavior and external traffic changes. Our integrated evaluation task thus requires MLLMs to synchronously interpret both streams and provide explanations grounded in visual evidence. This dual-view approach not only enhances the interpretability of driver understanding systems but also paves the way for safer and more transparent human-centered driving technologies.

\section{Methods} \label{sec:method}
\subsection{Pipeline Overview} \label{sec:pipeline}

Our dataset creation pipeline is illustrated in Figure \ref{fig:framework}, where we identify five data processing steps as follows:
\begin{figure*}
    \centering
    \includegraphics[width=.90\linewidth]{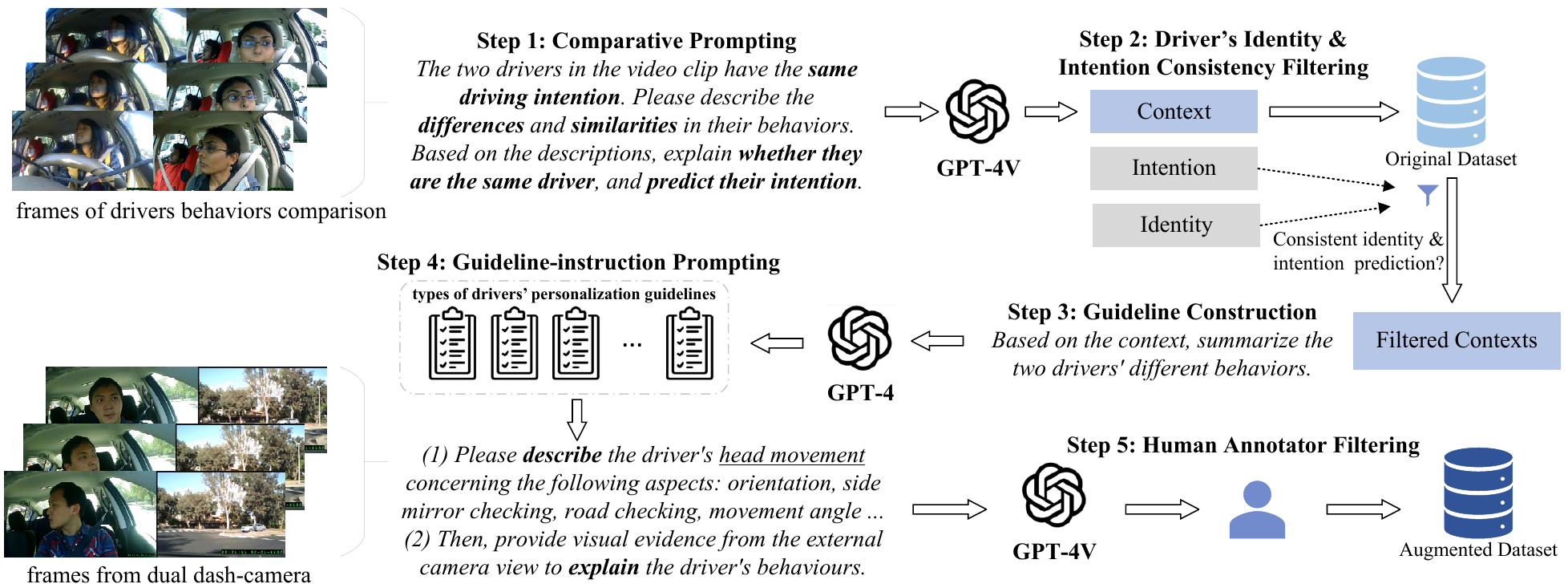}
    \caption{Illustration of the pipeline for creating a dataset that captures drivers' personalized behavior characteristics, including five key steps: 
    comparative prompting, consistency-based sample filtering, summarizing into generation guidelines, fine-grained behavior generation, and human annotator filtering for quality assurance}
    \label{fig:framework}
    \vspace{-.5cm}
\end{figure*} 
\begin{itemize}
    \item \textbf{Step 1 Comparative Prompting}: To understand personalized driver behavior, we propose comparative prompting in MLLMs for descriptions of driver behavior discrepancies while performing the same action.
    \item \textbf{Step 2 Driver’s Identity \& Intention Consistency Filtering}: We extract the intention and driver identities from the generated descriptions as validation measurements for automatic sample filtering.
    \item \textbf{Step 3 Guideline Construction}: Instead of directly using the extracted descriptions from comparative prompting, 
                           we extract driving behavior types and categorize descriptions as prompting guidelines, which are post-processed by text-only LLMs.
    \item \textbf{Step 4 Guideline-instruction Prompting}: By prompting MLLMs with the generation guidelines \cite{sainz2023gollie}, we can focus more effectively on specific driver behavior characteristics.
                           This approach leads to more fine-grained and personalized driver behavior, which reduces the chance of hallucinations.
    \item \textbf{Step 5 Human Annotator Filtering}: However, MLLMs may still hallucinate non-existing visual contexts in the limited visual field of internal and external dash-cams.
                           Therefore, the human annotation involves aspect-level and sample-level filtering for invisible and irrelevant behavior description detection. 
\end{itemize}
With the proposed pipeline, we augment dual-cam videos from the Brain4Cars \cite{jain2016brain4cars} dataset,
with personalized driver behavior descriptions and explanations.
The original Brain4Cars dataset was collected from 10 different drivers and segmented into 700 event-based clips with 274 lane changes, 131 turns, and 295 driving straight instances \cite{jain2016brain4cars}.

\subsection{Comparative Prompting}
% (Junda): Add abalation (compare more than 2 drivers)
% (Junda): Selectively filter (based on the data quality)
To obtain drivers' personalized behavior, we propose comparative prompting in MLLMs to extract behavior discrepancies between two drivers with the same intention.
For each pair of drivers $u,v$ with the same intention, we first extract $N$ frames from internal dash-cam video frames $\mathbf{I}_u^{\mathit{in}}, \mathbf{I}_v^{\mathit{in}} \in V_i$ as the visual evidence.
To construct a compatible visual context for MLLM prompting, temporal frames of drivers' videos are concatenated (\emph{e.g.}, Step 1 of Figure \ref{fig:framework}),
\begin{equation*}
    \mathbf{I}_{u,v}^{\mathit{in}} = \text{vconcat}\left[\text{hconcat}(\mathbf{I}_u^{\mathit{in}}), \text{hconcat}(\mathbf{I}_v^{\mathit{in}})\right],
\end{equation*}
in which \text{vconcat} and \text{hconcat} denote vertical and horizontal concatenations of image frames, respectively. 

Since MLLMs are developed from LLMs with multimodality alignment, MLLMs can have strong textual priors (\emph{i.e.}, linguistic bias) \cite{ko2023large}, 
which may lead to hallucinated responses neglecting visual evidence \cite{ko2023large,alayrac2022flamingo}.
Specifically, previous works point out that failing to instruct the models to focus on fine-grained aspects of visual semantics can result in hallucinations \cite{liu2024survey}.
Thus, in this step, we propose to prompt the MLLMs to generate only comparative descriptions between drivers, 
which make MLLMs perceive more fine-grained visual details \cite{zhang2024cocot} and potentially prevent hallucination problems \cite{liu2024survey,zhou2023analyzing}. 

The comparative prompt design in our pipeline consists of 4 main instructions. 
The \textbf{explanation} instruction $\mathbf{T}_x$ is to explicitly reason on the sequence of concatenated video frames as videos.
Then, the MLLM is prompted with the \textbf{comparison} instruction $\mathbf{T}_c$ to answer the discrepancies between the two drivers' behavior. 
To derive the driver's identity and intention consistency indexes, we prompt the MLLM with both \textbf{identity} $\mathbf{T}_{id}$ and \textbf{intention} $\mathbf{T}_{it}$ extraction instructions based on the model's previous comparative descriptions.
By the driver's identity and intention consistency evaluation, we can improve the accuracy of the MLLM's responses. 
% \begin{itemize}
%     \item \textbf{Explanation}: \textit{These are consecutive frames extracted from two different video. In each frame, two drivers are performing the same operation on their cars. Please answer the following questions: }
%     \begin{itemize}
%         \item \textbf{Comparison}: \textit{Please compare their behaviours and describe the similarities and differences between them, While they are performing the same operation on their cars.} 
%         \item \textbf{Identity}: \textit{Are the two videos containing the same driver or the different drivers? Explain your answer step by step according to visual evidence. }
%         \item \textbf{Intention}: \textit{Finally based on the behaviour analysis, predict their common action (options: go straight, turn left, turn right, turn to the left lane, turn to the right lane) and explain step by step.}
%     \end{itemize}
% \end{itemize}
Together with the visual evidence and textual instructions, we prompt the MLLM to generate initial responses,
% (detailed prompts are in Supplementary Material)
\begin{equation*}
    \mathbf{R}_{u,v} = \text{MLLM}\left(\mathbf{I}_{u,v}^{\mathit{in}}, [\mathbf{T}_x;\mathbf{T}_c;\mathbf{T}_{id};\mathbf{T}_{it}]\right),
\end{equation*}
in which we make sure the drivers' intentions are the same.

\subsection{Identity and Intention Consistency Filtering}
We introduce an intermediate sample verification and filtering step based on the driver's identity and intention consistency. 
The \textbf{context} instruction $\mathbf{T}'_c$ is to reformat the comparative descriptions into drivers' personalized behavior types and aspects into dictionaries, 
which leverages the LLM's structural text generation capacities.
We also prompt the textual LLM to extract the \textbf{identity} and \textbf{intention} predictions 
following the instructions $\mathbf{T}'_{id}$ and $\mathbf{T}'_{it}$ respectively.
% (detailed prompts are described in Supplementary Material).
% \begin{itemize}
%     \item \textbf{Context}: \textit{Based on the context, summarize the two drivers' different behaviours as a dictionary. 
%     In the dictionary, the type of behaviours is the key and their different behaviours are the values.}
%     \item \textbf{Identity}: \textit{Based on the context, summarize the conclusion whether the two drivers are of the same identity or not. }
%     % Brace the final answer between the special tokens "<" and ">" by answer either "same" or "different".
%     \item \textbf{Intention}: \textit{Based on the context, summarize the conclusion of the two drivers' intentions. }
%     % Brace the final answer between the special tokens "<" and ">" by answer in the following options: turn right, turn left, change right, change left, remain straight.
% \end{itemize}
By prompting the textual LLM with instructions and the previous response $\mathbf{R}_{u,v}$, we can extract the personalization context $\mathbf{C}_{u,v}$, 
drivers' identity indicators $\mathbf{1}^{\mathit{ID}}_{u,v}$, and intentions $\mathbf{IT}_{u}, \mathbf{IT}_{v}$ respectively.
% \begin{align*}
%      \mathbf{C}_{u,v} &= \text{LLM}\left(\mathbf{R}_{u,v}, \mathbf{T}'_c\right), \\
%      \mathbf{1}^{\mathit{ID}}_{u,v} &= \text{LLM}\left(\mathbf{R}_{u,v}, \mathbf{T}'_{id}\right), \\
%      \mathbf{IT}_{u}, \mathbf{IT}_{v} &= \text{LLM}\left(\mathbf{R}_{u,v}, \mathbf{T}'_{it}\right).
% \end{align*}
Then, the personalization contexts are filtered and aggregated into $P$ according to the identity and intention consistency, 
% $P =  \bigcup_{\substack{\mathbf{I}_u, \mathbf{I}_v \in V_i, \\ \mathbf{1}^{\mathit{ID}}_{u,v}=\text{ID}(u,v), \\ \mathbf{IT}_{u}=\mathbf{IT}_{v}}}  \mathbf{C}_{u,v}$
\begin{equation*}
   P =  \bigcup_{\substack{\mathbf{I}_u, \mathbf{I}_v \in V_i, \\ \mathbf{1}^{\mathit{ID}}_{u,v}=\text{ID}(u,v), \mathbf{IT}_{u}=\mathbf{IT}_{v}}}  \mathbf{C}_{u,v},
\end{equation*}
in which the identity is to check if the generated identity-matching result $\mathbf{1}^{\mathit{ID}}_{u,v}$ is the same as the ground truth identities $\text{ID}(u,v)$.
Since comparative prompting compares two drivers with the same intention, we validate if the generated intentions $\mathbf{IT}_{u}, \mathbf{IT}_{v}$ are also the same.

\subsection{Guideline Instruction Construction and Prompting}
The MLLM can generate detailed contrastive characteristics in drivers' behavior by comparative prompting. 
However, in practice, we observe hallucinations in descriptions of visual details. 
Due to lacking granularity in visual understanding, the MLLM can respond with visual details that do not exist in the actual visual context \cite{zhang2024cocot,liu2024survey,zhou2023analyzing}.
To make the prompt specific in describing a certain type of driver's behavior and prominent visual characteristics of the behavior, 
we propose aggregating the collected personalization contexts $D$ as the prompting guideline instructions.
Following the guideline instructions for each type of personalized behavior, the generated responses contain less irrelevant information and hallucinations \cite{sainz2023gollie}.

\begin{figure*}
    \centering
    \includegraphics[width=.90\linewidth]{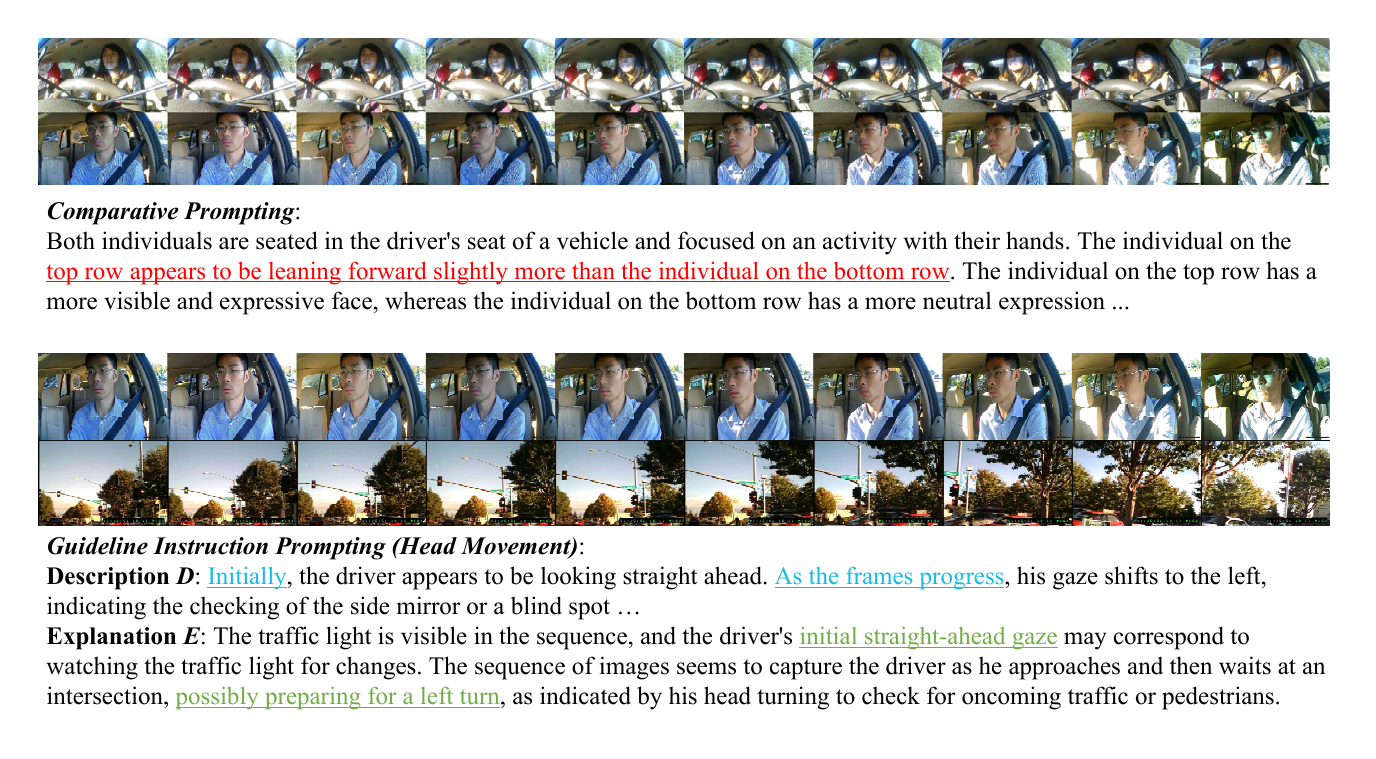}
    \caption{A comparative example of comparative prompted description and guideline instruction prompted description. 
    The context (underlined and in red) indicates possible hallucinations by \textbf{comparative prompting}. 
    With fine-grained guideline instructions, more detailed visual evidence and temporal information (underlined and in blue) can be prompted from LMMs in \textbf{Description \textit{D}}.
    Then, the LMMs can capture visual evidence from an external view and explain the driver's behavior (underlined and in green) in \textbf{Explanation \textit{E}}.}
    \label{fig:guideline_eg}
    \vspace{-.5cm}
\end{figure*}
The constructed guideline instructions are types of personalized driver behavior paired with detailed aspects of these types of behavior (\emph{e.g.}, in Step 4 of Figure \ref{fig:framework}).
During guideline instruction prompting, the MLLM is individually prompted with each type of behavior with its descriptions, to prevent irrelevant information and hallucinations.
In addition, we also prompt the MLLM to explain the driver's behavior description based on visual evidence from the external dash-cam frames (\emph{e.g.}, in Step 4 of Figure \ref{fig:framework}). 
To enable the MLLM to perceive both the internal and external visual evidence based on video frames, 
we use a similar way to construct the visual evidence.
Given a driver's internal $\mathbf{I}_u^{\mathit{in}}\in V_i$ and external $\mathbf{I}^{\mathit{ex}}_{u}\in V_e$ dash-cam video frames, 
\begin{equation*} \label{eq:dual-cam}
    \mathbf{I}_{u}^{\mathit{du}} = \text{vconcat}\left[\text{hconcat}(\mathbf{I}_u^{\mathit{in}}), \text{hconcat}(\mathbf{I}^{\mathit{ex}}_{u})\right].
\end{equation*}
For each behavior type $k\in K$ and its description $S(k)$, the prompt $T_{g}(k,S(k))$ is the guideline instruction for prompting.
% (detailed prompts are described in Supplementary Material).
% \begin{itemize}
%     \item \textbf{Description}: \textit{Please consider the driver's $<k>$ concerning the following aspects $<S(k)>$, and output a summary in detail.}
%     \item \textbf{Explanation}: \textit{Then, provide visual evidence from the external camera view to justify the driver's behaviours that you summarized. }
% \end{itemize}
With the guideline instructions, we can prompt the MLLM to obtain the descriptions $\mathbf{D}_{u}$ and explanations $\mathbf{E}_{u}$,
\begin{equation*}
    \mathbf{D}_{u}, \mathbf{E}_{u} = \text{MLLM}\left(\mathbf{I}_{u}^{\mathit{du}}, T_{g}(k,S(k))\right).
\end{equation*}

In Figure \ref{fig:guideline_eg}, we show a comparative example of the difference between the description prompted from comparative prompting and guideline instruction prompting.
% In comparative prompting, the MLLM is prompted to describe the general behavior discrepancies between two drivers.
In comparative prompting, we can observe only coarse-grained aspects (\emph{e.g.}, seating position, facial expression,  etc.) are mentioned without further justification.
In such descriptions, we can observe possible hallucinations generated from MLLMs (\emph{e.g.}, the first driver is leaning forward more than the other driver).
In the guideline instruction prompted responses, we can observe more fine-grained descriptions about the driver's behavior changing with time, which is also specific in a certain aspect without irrelevant information.
The MLLM can further try to explain the generated description based on corresponding visual evidence from the external dash-cam frames.

\subsection{Human Annotator Filtering}
% Human annotators are involved in sample filtering, to reduce human effort and improve data creation scalability.
One of the sample filtering tasks is to detect irrelevant or inaccurate types of behavior, namely aspect-level sample filtering.
We observe several extracted behaviors (\emph{e.g.}, feet position, leg movement) that cannot be accurately described due to limited visual field from the internal dash-cam. 
In such cases, the MLLM will hallucinate inaccurate visual descriptions and provide false explanations for the descriptions.
In addition, some other extracted behaviors (\emph{e.g.}, driver's appearance, attire, etc.) are irrelevant to the performance of driving tasks.
In such cases, the explanations are not based on visual evidence but on MLLMs' strong textual priors (\emph{i.e.}, linguistic bias) \cite{ko2023large}.
% although MLLMs can accurately describe these aspects of the drivers, the explanations are potential hallucinations from the MLLM based on its biased assumptions.

After aspect-level filtering, we can obtain nine types of behavior in Table \ref{tab:types}.
The sample-level filtering task for human annotators is mainly to filter out failure cases from MLLMs.
Since these failure cases of generation follow a similar pattern of expression (\emph{e.g.}, ``I'm sorry, but I cannot provide ...''), 
including such samples in model fine-tuning may result in overfitting problems. Thus, we further conduct the sample-level inspection of these specific failure cases.

\begin{table}
    \small
    \centering
    \caption{Behavior types after the human annotator's filtering.}
    \label{tab:types}
    \begin{tabular}{ll|ll}
        \toprule
        \textbf{ACT} & action   & \textbf{BOL} & body language  \\
        \textbf{DRS} & driving style  & \textbf{FAE} & facial expression \\
        \textbf{GEM} & gaze/eye movement & \textbf{HAM} & hand movement      \\
        \textbf{HEM} & head movement     & \textbf{INT} & intention    \\
        \textbf{IWP} & interaction w/ passenger \\ 
        \bottomrule
    \end{tabular}
    \vspace{-1em}
\end{table}
\begin{table*}
\small
\centering
\caption{
% We report the statistics of the training and test datasets. 
The number of clips, description-explanation pairs (D/E), question-answering pairs (QA), average words in descriptions (Desc.), explanations (Expl.), questions (Q), and answers (A).
}
\label{table:stat}
\begin{tabular}{c|ccc|cccc}
\toprule
               & \multicolumn{3}{c|}{Sample Number} & \multicolumn{4}{c}{Average Length}    \\
               \cmidrule(lr){2-4}                      \cmidrule(lr){5-8}  
               & \textbf{ Clips} & \textbf{ D/E} & \textbf{ QA} & \textbf{Desc.} & \textbf{Expl.} & \textbf{Q} & \textbf{A} \\
\toprule
\textbf{Train} &  478    & 5,084            & 35,972          & 125.06          & 100.45          & 13.14       & 37.69      \\
\textbf{Test}  & 116               & 1,224            & 8,881           & 126.20           & 100.27          & 13.13       & 37.59       \\ 
\bottomrule
\end{tabular}
\end{table*}
\section{Personalized Driving Behavior Evaluation (PDB-Eval)}
\begin{figure*}
    \centering
    \includegraphics[width=.90\linewidth]{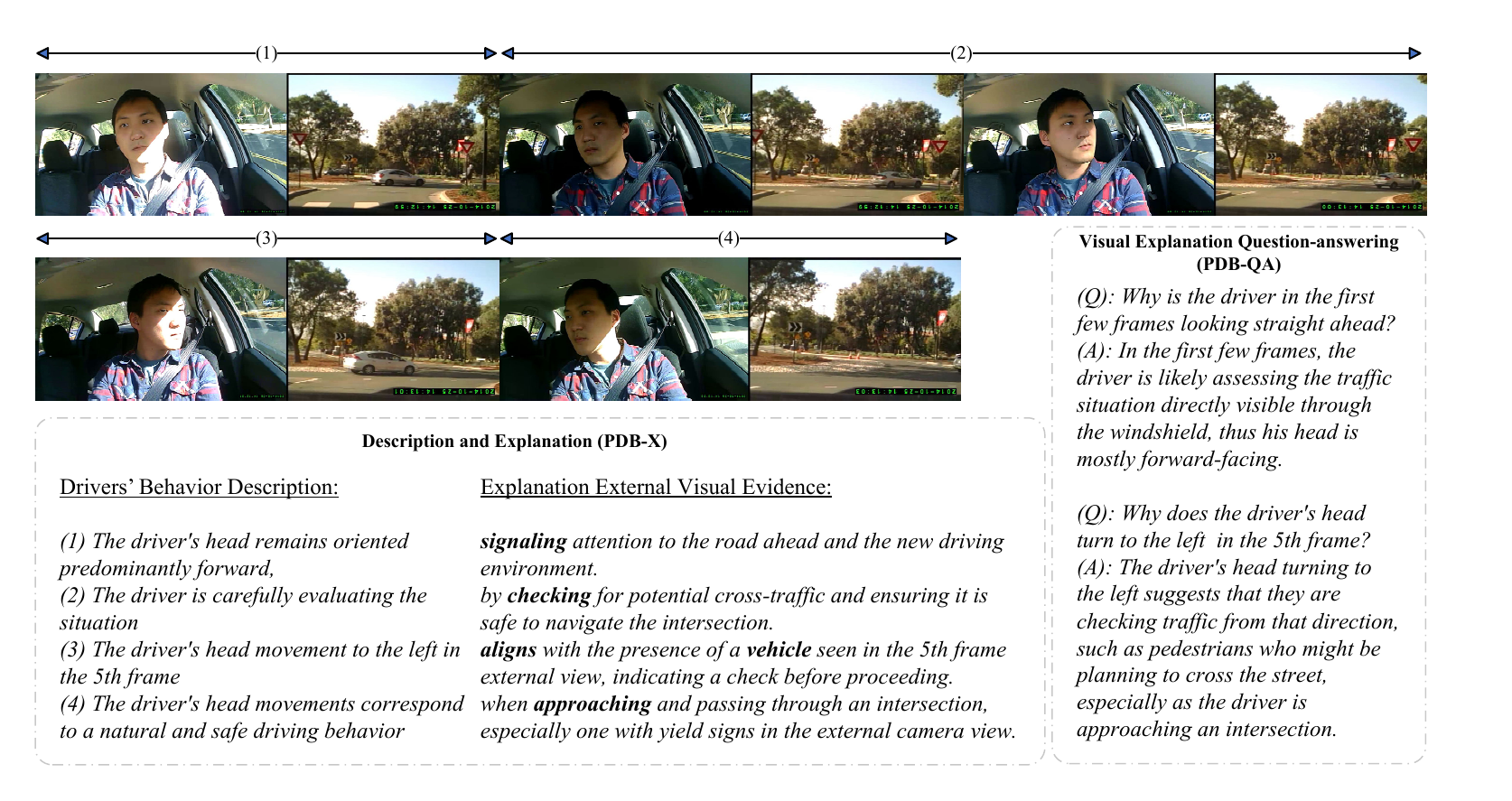}
    \caption{Illustration of three tasks for drivers' personalized behaviors understanding.}
    \label{fig:dataset_eg}
    % \vspace{-.5cm}
\end{figure*}

Based on our data creation pipeline in Section \ref{sec:pipeline}, we develop our evaluation dataset from the existing Brain4Cars dataset \cite{jain2016brain4cars}.
% for understanding drivers' personalized behavior for three main tasks: description, explanation, and question-answering.
In Figure \ref{fig:dataset_eg}, we illustrate an example in our constructed dataset for these three tasks. 
In comparative prompting, we sampled 20 pairs of drivers from each annotated driving intention subset: right turn, left turn, right change, left change, and driving straight.
Through comparative prompting, the collected comparative descriptions are summarized into \textbf{nine types} in Table \ref{tab:types} with an average of \textbf{19.11 guideline instructions} for each type.
We summarize the statistics of PDB-Eval in Table \ref{table:stat}, where we report the number of clips, description-explanation pairs (D/E), and question-answering pairs (QA). 
We also report the average number of words in descriptions (Desc.), explanations (Expl.), questions (Q), and answers (A).

In the PDB-X dataset, we observe that the average length of the descriptions is longer than that of the explanations, which is because of the usage of the guideline instructions.
Based on each requirement in the guideline, the MLLM will generate an average of \textbf{6.54 words per instruction} for the description.
As for the PDB-QA dataset, the average number of questions derived from the PDB-X dataset is \textbf{7.05 QA pairs per D/E pair}, 
and the average length of the ground-truth answers is around \textbf{3 times the length} of questions.
% To further visually illustrate the generated descriptions, explanations, and question-answering pairs, 
% we show keyword frequencies in Figure \ref{fig:wc} for descriptions (Figure \ref{fig:desc_wc}) and explanations (Figure \ref{fig:expl_wc}),
% which represent the word frequencies of the generated contexts.
% In Figure \ref{fig:desc_wc}, we can find that the most frequent keywords in descriptions (\emph{e.g.}, head movement, hand position, etc.) are aligned with the summarized types of behavior in Table \ref{tab:types}.
% In Figure \ref{fig:expl_wc}, we can observe that the keywords in explanations (\emph{e.g.}, traffic, intersection, turn sign, etc) describing the external traffic situation have a relatively higher frequency.
% \input{figures/desc}
% Based on the PDB-X dataset, we develop a visual explanation question-answering task PDB-QA,
% which is more compatible with general pipelines of MLLM instruction fine-tuning \cite{rajpurkar2016squad, lewis2019mlqa, brown2020language}
Based on the descriptions and explanations from PDB-X, we can enable automatic question-answer pair generation using text-only LLMs \cite{guo2023images,kim2024generalizing} 

\section{Experiments}
We evaluate various MLLMs: BLIP-2 \cite{li2023blip}, VTimeLLM \cite{huang2023vtimellm}, and GPT-4V, on PDB-X and PDB-QA.
By fine-tuning (BLIP-2 and VTimeLLM) on PDB-X, we further evaluate the MLLMs on downstream driving benchmarks, Brain4Cars \cite{jain2016brain4cars} (in-domain) and AIDE \cite{yang2023aide} (cross-domain),
to understand the effectiveness of the generated drivers' personalized behavior descriptions and explanations. 
% The implementation and training details about MLLMs fine-tuning are included in \textbf{Supplementary Material}.

\subsection{Data Preprocessing} \label{sec:data-preprocess}
We adopt different video preprocessing methods for three baselines to extract compatible visual representations of the input streams.
For BLIP-2, we extract 10 image frames with an equal time interval from dual-cam and concatenate them as in Eq. \eqref{eq:dual-cam} before encoding the image using 
the CLIP encoder from BLIP-2 \cite{li2023blip}.
For VTimeLLM, we follow the preprocessing method in \cite{huang2023vtimellm} to extract 10 image frames from each second of the video and encode the video with 
a ViT-based CLIP model.
For GPT-4V, we extract 10 consecutive image frames from the videos and concatenate each frame of the internal and external videos together.

\subsection{Description and Explanation (PDB-X)}
We evaluate the MLLM baselines for description and explanation tasks in PDB-X and report the BLEU-4 metric of comparative results in Table \ref{table:exp}.
For open-source MLLMs (BLIP-2 and VTimeLLM) we first fine-tune the models on the training set of PDB-X before evaluation.
During inference, we adopt the same instruction design and visual encoding as in the fine-tuning stage.
Comparing the performance of fine-tuned MLLMs and GPT-4V, we can observe a consistent advantage of fine-tuning on PDB-X.
Such an observation suggests that prompting MLLMs with simple instructions cannot directly enable fine-grained descriptions and explanations,
which showcases the effectiveness of the proposed comparative prompting method and the challenges posed by PDB-X tasks. 
\begin{table}[ht]
\small
\centering
\caption{
Performance of \textit{BLEU-4} metrics on all types of driver behavior (in Table \ref{tab:types}) and the two tasks, description (Desc.) and explanation (Expl.). 
We indicate the best performance for the description and explanation tasks in bold.
}
\label{table:exp}
\begin{tabular}{c|cc|cc|cc}
\toprule
                           & \multicolumn{2}{c|}{\textbf{BLIP-2}} & \multicolumn{2}{c|}{\textbf{VTimeLLM}} & \multicolumn{2}{c}{\textbf{GPT-4V}} \\
                           \cmidrule(lr){2-3}                      \cmidrule(lr){4-5}    \cmidrule(lr){6-7}  
Type           & Desc.  & Expl.  & Desc.   & Expl.  & Desc.  & Expl.  \\ 
\midrule
\textbf{ACT} & \textbf{51.10}    & 48.09 & 47.18             & \textbf{50.41}    & 29.49             & 28.57 \\
\textbf{BOL} & 34.51 & 52.74     & 49.13    & \textbf{62.03}    & 33.68             & 27.22 \\
\textbf{DRS} & 72.14             & \textbf{58.50}    & \textbf{86.03}    & 54.63             & 32.33             & 30.12 \\
\textbf{FAE} & 54.52             & \textbf{59.90}    & \textbf{72.17}    & 50.40             & 34.44             & 28.54 \\
\textbf{GEM} & 55.06             & 57.83             & \textbf{62.72}    & \textbf{60.70}    & 39.22             & 35.73 \\
\textbf{HAM} & \textbf{46.42}    & \textbf{70.50}    & 42.93 & 44.98 & 40.96             & 36.54 \\
\textbf{HEM} & 72.20             & \textbf{51.94}    & \textbf{72.90}    & 46.71             & 39.87             & 36.77 \\
\textbf{INT} & \textbf{56.52}    & \textbf{60.97}    & 53.80             & 49.22             & 25.69 & 26.67 \\
\textbf{IWP} & \textbf{76.39}    & \textbf{66.53}    & 73.80             & 65.95             & 37.23             & 32.04 \\ \hline
\textbf{AVE} & 57.65             & \textbf{58.55}    & \textbf{62.30}    & 53.89             & 34.77             & 31.35 \\
\bottomrule
\end{tabular}
\end{table}

Comparing the performance of BLIP-2 and VTimeLLM, we can observe that these two MLLMs are specialized in the explanation and description tasks respectively.
We argue that BLIP-2 is pre-trained with enhanced visual knowledge reasoning capacity \cite{li2023blip} that benefits visual explanation, 
while the static modality of the model input limits its description capacity.
On the other hand, VTimeLLM is specially pre-trained with the temporally-aware descriptive capacity for fine-grained video moment understanding \cite{huang2023vtimellm},
which explains its better performance on the description task.
Nevertheless, creating an MLLM that achieves robust performance in both tasks remains challenging.

\subsection{Visual Explanation QA (PDB-QA)}
We further evaluate MLLMs on PDB-QA to showcase their capacities in answering complex questions related to personalized driving behavior.
% According to the question distribution in Figure \ref{fig:qa_hier}, most of the questions in PDB-QA are about personalization evidence or justifications for the driver's behavior.
The relatively lower performance on PDB-QA suggests a deficiency of the MLLMs in understanding drivers' personalized information.
However, with model fine-tuning, we can still observe consistent improvements in both open-source MLLMs.
\begin{table}[ht]
\small
% \small
\centering
\caption{
Performance of \textit{BLEU-4} metrics on the PDB-QA dataset, including pre-trained (PT), fine-tuned (FT), and zero-shot (ZS). 
% The PT performance is obtained from models only pre-trained on PDB-X tasks. 
% The FT performance is obtained from models that are further fine-tuned on the training set of PDB-QA.
% The ZS performance is obtained from GPT-4V which is zero-shot prompted.
}
\label{table:qa}
\begin{tabular}{c|cc|cc|c}
\toprule
                  & \multicolumn{2}{c|}{~\textbf{BLIP-2}~} & \multicolumn{2}{c|}{~\textbf{VTimeLLM}~} & \textbf{GPT-4V}   \\
                           \cmidrule(lr){2-3}                      \cmidrule(lr){4-5}    \cmidrule(lr){6-6}  
~Metric~           & ~PT $\uparrow$~ & ~FT $\uparrow$~ & ~PT $\uparrow$~ & ~FT $\uparrow$~ & ~ZS $\uparrow$~  \\ 
\midrule
~\textbf{BLEU-4}~ & 30.29 & 33.64 & 28.07 & \textbf{48.61} & 16.08  \\
\bottomrule
\end{tabular}
\end{table}

Comparing the performance of BLIP-2 and VTimeLLM, we can observe that VTimeLLM has a larger fine-tuning improvement compared to its zero-shot performance, 
which suggests that VTimeLLM has a higher performance upper-bound than BLIP-2.
The temporal reasoning capacity in VTimeLLM can enhance the video moment understanding between a certain time boundary \cite{huang2023vtimellm}.
However, how to enable cross-validation of the reasoning in video clips within different time boundaries for better personalization understanding can still be challenging,
which requires both temporal understanding and causal reasoning abilities.

\subsection{Driving Task Evaluation}
In this section, we evaluate the effectiveness of visual descriptions and explanations in several driving tasks,
with the fine-tuned MLLMs' outputs as textual evidence and the corresponding video features as visual evidence. 
For evaluation consistency, we use the same video features described in Section \ref{sec:data-preprocess} for each MLLM
and textual evidence from MLLMs' fine-tuned on PDB-X.

\subsubsection{Evaluation Results on Brain4Cars \cite{jain2016brain4cars}.}
We first evaluate the intention prediction task proposed in the Brain4Cars \cite{jain2016brain4cars} dataset as the \textbf{in-domain} evaluation.
To demonstrate the generalizability of MLLMs, we do not incorporate additional low-level signals (\emph{e.g.}, vehicle speed, and GPS information) used by Brain4Cars.   

In Table \ref{table:intent}, we  observe that drivers' personalized behavior can be good complementary textual evidence for prediction of drivers' turn maneuvers,
given that these maneuvers typically exhibit more discernible behavior from the driver.
On the other hand, the original method in Brain4Cars, which fuses various sensory information sources, still outperforms MLLMs with only vision-language information in lane change prediction.
We argue that in tasks requiring additional sensory information, our method may still incorporate the information into the general reasoning process.
\begin{table}[ht]
\small
\centering
% \begin{tabular}{l|cc|cc|cc|cc}
% \toprule
%                      & \multicolumn{2}{c|}{\textbf{BLIP-2}} & \multicolumn{2}{c|}{\textbf{VTimeLLM}} & \multicolumn{2}{c|}{\textbf{S-RNN}} & \multicolumn{2}{c}{\textbf{PDB-X}} \\
%                            \cmidrule(lr){2-3}                      \cmidrule(lr){4-5}    \cmidrule(lr){6-7}  \cmidrule(lr){8-9}  
%                      & Prec.           & Recall         & Prec.            & Recall          & Prec.          & Recall          & Prec.          & Recall        \\ 
% \midrule
% \textbf{Turn}        & 84.57              & 76.43             & 83.85               & 77.18              & 75.20               & 75.30                & \textbf{87.24} & \textbf{81.61}         \\
% \textbf{Lane Change} & 76.47              & 72.73             & 76.04               & 72.11              & \textbf{85.40}      & \textbf{86.00}       & 76.04          & 72.17         \\ 
\caption{
Performance of precision and recall metrics on the intention prediction task \cite{jain2016brain4cars}. 
The BLIP-2 and VTimeLLM columns are obtained with textual evidence output from LMMs fine-tuned on PDB-X.
% The PDB-X column is obtained with the original textual evidence in PDB-X, which serves as the upper bound of our method.
The S-RNN column is obtained from the method from the original paper \cite{jain2016brain4cars}.
}
\label{table:intent}
\begin{tabular}{l|cc|cc|cc}
\toprule
                     & \multicolumn{2}{c|}{\textbf{BLIP-2}} & \multicolumn{2}{c|}{\textbf{VTimeLLM}} & \multicolumn{2}{c}{\textbf{S-RNN}} \\
                           \cmidrule(lr){2-3}                      \cmidrule(lr){4-5}    \cmidrule(lr){6-7}  
                     & Prec.           & Recall         & Prec.            & Recall          & Prec.          & Recall        \\ 
\midrule
\textbf{Turn}        & \textbf{84.57}              & 76.43             & 83.85               & \textbf{77.18}              & 75.20               & 75.30                      \\
\textbf{Change}      & 76.47              & 72.73             & 76.04               & 72.11              & \textbf{85.40}      & \textbf{86.00}           \\ 
\bottomrule
\end{tabular}
\end{table}

\subsubsection{Evaluation Results on AIDE \cite{yang2023aide}.}
We further evaluate the MLLMs that are initially fine-tuned on PDB-X without additional fine-tuning on AIDE as the \textbf{cross-domain} evaluation. 
The tasks proposed by AIDE are driver behavior recognition (DBR), driver emotion recognition (DER), traffic context recognition (TCR), and vehicle condition recognition (VCR).
In addition to the dual-cam videos used in our method, AIDE \cite{yang2023aide} additionally incorporates videos from the left and right cameras, and multimodal annotations of a driver's face, body, gesture, and posture.
\begin{table}[ht]
% \vspace{-.5cm}
\small
\centering
\caption{
Performance of Acc and weighted F1 metrics on the AIDE dataset \cite{yang2023aide}. 
The BLIP-2 and VTimeLLM columns are obtained by the LMMs that are initially fine-tuned on PDB-X and then used for AIDE without additional fine-tuning.
}
\label{table:aide}
\begin{tabular}{c|cc|cc|cc}
\toprule
                  & \multicolumn{2}{c|}{\textbf{BLIP-2}} & \multicolumn{2}{c|}{\textbf{VTimeLLM}} & \multicolumn{2}{c}{\textbf{AIDE}}    \\
                           \cmidrule(lr){2-3}                      \cmidrule(lr){4-5}    \cmidrule(lr){6-7}  
                  & Acc $\uparrow$ & F1 $\uparrow$ & Acc $\uparrow$ & F1 $\uparrow$ & Acc $\uparrow$ & F1 $\uparrow$  \\ 
\midrule
\textbf{DER} & 72.74 & 72.73 & \textbf{74.06} & \textbf{73.41} & 71.26 & 68.71 \\
\textbf{DBR} & 64.86 & 64.41 & \textbf{65.52} & \textbf{65.34} & 65.35 & 63.29 \\
\textbf{TCR} & \textbf{90.15} & 89.86 & \textbf{90.15} & \textbf{90.19} & 83.74 & 81.28 \\
\textbf{VCR} & 76.35 & 76.00 & \textbf{78.82} & \textbf{77.67} & 77.12 & 75.23 \\
\bottomrule
\end{tabular}
\end{table}

In Table \ref{table:aide}, we can observe robust generalizability of the fine-tuned MLLMs in cross-domain driving tasks,
where the fine-tuned VTimeLLM achieves better accuracy and weighted F1 than AIDE in all tasks.
In the four driving tasks in AIDE, we observe better improvements in driver emotion recognition (DER) and traffic condition recognition (TCR), 
which benefit from the drivers' personalized behavior descriptions and external visual explanations respectively.

\section{Conclusion}
% We establish the Driver's Personalized Behavior Evaluation dataset (PDB-Eval), for drivers' personalized behavior analysis, leveraging both in-cabin and external traffic views. 
% The two components of our benchmark, PDB-X and PDB-QA, are obtained from the proposed visual comparative prompting within MLLMs,
% which captures behavior discrepancies in drivers, while describing and explaining based on the dual-cam viewpoints.
% By model fine-tuning on various MLLMs, we demonstrate consistent improvements on PDB-X and PDB-QA compared with the zero-shot performance of GPT-4V by up to 73.2\%.
% We assess the fine-tuned MLLMs in Brain4Cars' intention prediction and AIDE's recognition tasks, 
% showcasing their enhanced accuracy by up to 12.5\% and 11.0\% in driving-related prediction and recognition tasks of Brain4Cars and AIDE respectively.

% However, the relatively low BLEU-4 performance of the MLLMs also suggests remaining challenges for more accurate visual understanding in driving.
% Due to the inherent challenge of hallucination in MLLMs, more fine-grained multimodal reasoning over multi-view video streams can be challenging for existing MLLMs.
% We suggest future explorations to enable MLLMs with capacities to identify localized visual evidence with temporal awareness, which serves as intermediate chain-of-thought visual reasoning steps.
% Comparing MLLMs with and without temporal understanding abilities, we also emphasize the importance of temporal and causal reasoning in driving understanding.
We introduce the Driver’s Personalized Behavior Evaluation dataset (PDB-Eval), which leverages in-cabin and external views for personalized driver behavior analysis. Our benchmark comprises two components—PDB-X and PDB-QA—derived via visual comparative prompting within MLLMs to capture behavioral discrepancies and provide descriptive explanations. 
Fine-tuning various MLLMs improved performance on these tasks by up to 73.2\% over GPT-4V’s zero-shot results.
% and enhanced accuracy in Brain4Cars' intention prediction and AIDE's recognition by up to 12.5\% and 11.0\%, respectively. 
However, the relatively low BLEU-4 scores indicate persistent challenges in fine-grained, temporally aware multimodal reasoning. 
Future work should focus on enabling MLLMs to identify localized visual evidence with temporal awareness.

\bibliographystyle{IEEEtran}
\bibliography{egbib}

% \vspace{12pt}
% \color{red}
% IEEE conference templates contain guidance text for composing and formatting conference papers. Please ensure that all template text is removed from your conference paper prior to submission to the conference. Failure to remove the template text from your paper may result in your paper not being published.

\end{document}